\documentclass[conference]{IEEEtran}
\usepackage{times}

\usepackage[numbers]{natbib}
\usepackage{multicol}
\usepackage[bookmarks=true]{hyperref}

\usepackage{amsfonts}	
\usepackage{amsmath}	
\usepackage{amssymb}    
\usepackage{mathtools}
\usepackage{siunitx}

\usepackage{booktabs}
\usepackage{makecell}  
\usepackage[flushleft]{threeparttable}  
\usepackage{multirow}

\usepackage[table]{xcolor}
\usepackage{xcolor}    
\usepackage{textcomp}  
\usepackage{gensymb}   
\usepackage{graphicx} 
\usepackage{xspace}

\usepackage[capitalize]{cleveref} 

\usepackage{flushend}

\crefname{section}{Sec.}{Secs.}
\Crefname{section}{Section}{Sections}
\Crefname{table}{Table}{Tables}
\crefname{table}{Tab.}{Tabs.}

\definecolor{Gray}{gray}{0.9}

\begin{document}

\title{\Huge Efficient Robot Learning for Perception and Mapping}

\author{Author Name Omitted for Anonymous Review.}

\author{\authorblockN{Niclas Vödisch}
\authorblockA{University of Freiburg} \vspace*{-.6cm}}


%


\maketitle

\IEEEpeerreviewmaketitle


\section{Introduction}
Holistic scene understanding poses a fundamental contribution to the autonomous operation of a robotic agent in its environment. Key ingredients include a well-defined representation of the surroundings to capture its spatial structure as well as assigning semantic meaning while delineating individual objects.
Classic components from the toolbox of roboticists to address these tasks are simultaneous localization and mapping (SLAM) and panoptic segmentation.
Although recent methods demonstrate impressive advances, mostly due to employing deep learning, they commonly utilize in-domain training on large datasets. Since following such a paradigm substantially limits their real-world application, my research investigates \textbf{how to minimize human effort in deploying perception-based robotic systems to previously unseen environments}.
In particular, I focus on leveraging continual learning and reducing human annotations for efficient learning.

{\parskip=3pt
\noindent\textbf{Continual Learning}
(CL)~\cite{lopez2017gradient} and lifelong learning~\cite{thrun1995is} aim at overcoming the traditional approach that a learning-based model is trained for a specific task, defined a priori, with a fixed set of data. During inference, the model is then employed to previously unseen data from the same domain. However, in real-world applications, the data characteristics can change over time and differ between environments. Classic domain adaptation~\cite{guizilini2021geometric, lopez2020desc, tranheden2021dacs} attempts to bridge this gap between a source domain $\mathcal{S}$ and a target domain $\mathcal{T}$ but rarely considers maintaining the performance on the source domain, which can result in catastrophic forgetting. Thus, my research explores exploiting CL concepts for unsupervised robot learning.
}

{\parskip=3pt
\noindent\textbf{Label-Efficient Learning}
is more similar to the aforementioned classical learning scheme but attempts to significantly reduce the amount of human annotations. The key idea is to enable inexpensive in-domain training via various forms of weak supervision~\cite{shen2023asurvey}. With respect to image segmentation, previous methods rely on contrastive learning~\cite{gansbeke2021maskcontrast}, sparse supervision from point~\cite{li2023point2mask} or scribble~\cite{pan2021scribble} annotations, or semi-supervised techniques~\cite{hoyer2021three, yang2022stpp}. In contrast, my research is inspired by the recent success of vision foundation models~\cite{caron2021dino, kirillov2023sam, oquab2023dinov2} and investigates how to exploit their semantically rich image representations for training downstream tasks with almost zero labels.
}

\begin{figure*}[t]
    \centering
    \includegraphics[width=\linewidth]{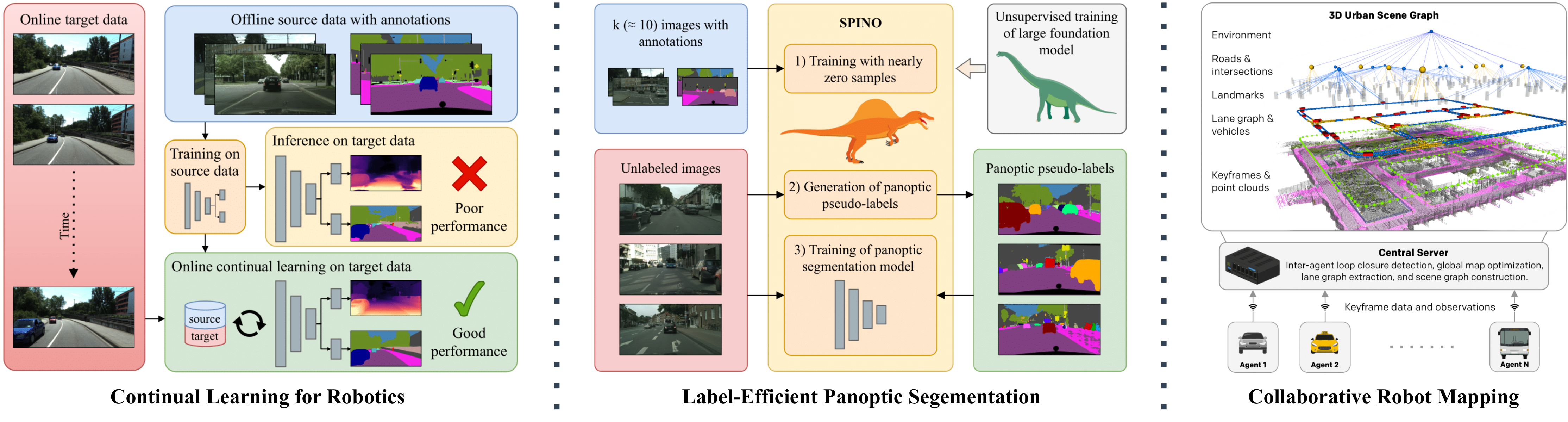}
    \vspace*{-1cm}
\end{figure*}

\section{Contributed Research}

Efficient robot learning for perception and mapping can be addressed by a variety of different paradigms. In the following, I summarize my previous works\footnote{To acknowledge the contributions of my co-authors, I utilize ``we''.} grouped by techniques.

{\parskip=3pt
\noindent\textbf{Continual Learning for Robotics}
equips an autonomous agent with the capability to automatically adapt to unseen domains while retaining high performance on all previous domains. We apply this idea to both SLAM and panoptic segmentation. While many previous works address sim-to-real adaptation~\cite{guizilini2021geometric}, we explicitly focus on transferring knowledge between real-world domains while alleviating catastrophic forgetting.
With respect to SLAM, we defined the novel concept of continual SLAM~\cite{voedisch2023clslam} that combines both lifelong SLAM, i.e., the long-term operation in a single spatially confined environment, and domain adaptation, i.e., the directed transfer of knowledge from a source to a target domain. In reality, continual SLAM implies that an autonomous vehicle can be trained in London and then employed in Delft, where it will adapt on the fly without human supervision but does not suffer from performance degradation when returning to London.
Addressing this task for vision-based SLAM, we utilize unsupervised depth estimation as an auxiliary task for joint online training with a separate odometry network. To balance adaptation to new environments and memory retention of preceding environments, we proposed a dual-network architecture and tackle forgetting by leveraging a replay buffer that stores raw images from previous domains that can be revisited in the network update steps. In a follow-up work~\cite{voedisch2023covio}, we proposed a diversity-driven sampling scheme that selectively updates this buffer when its size is limited, reflecting hardware constraints on robotic platforms.

Concurrently, we employed online continual learning for joint depth estimation and panoptic segmentation~\cite{voedisch23codeps}. In contrast to previous methods~\cite{guizilini2021geometric, lopez2020desc}, our approach can be adapted during deployment and specifically addresses forgetting. In this work, we proposed a novel cross-domain mixing strategy that combines labeled data from a source domain with unlabeled images from the online target domain to enable unsupervised adaptation of panoptic segmentation.

In summary, our research contributed to the continual unsupervised adaptation of robotic vision systems to new domains while not decreasing the performance in prior environments.
}

{\parskip=3pt
\noindent\textbf{Label-Efficient Panoptic Segmentation}
renders an important step to the widespread adoption of panoptic segmentation networks in robotic use cases. While recent methods~\cite{cheng2020panoptic, cheng2022mask2former} have shown great progress in terms of segmentation performance, they tend to rely on a vast amount of densely labeled training data whose generation is expensive and laborious~\cite{cordts2016cityscapes}. Recently, the vision community has demonstrated that task-agnostic pretraining yields image representations that can be bootstrapped for several downstream tasks~\cite{hamilton2022stego, oquab2023dinov2}. Therefore, we argue that it is time for a fundamental paradigm switch exploiting vision foundation models for extremely label-efficient training. We have shown that in contrast to unsupervised techniques~\cite{hamilton2022stego, hyun2021picie}, such an approach can yield results that are competitive with fully supervised learning methods.

In our research~\cite{kaeppeler2024spino, voedisch2024pastel}, we exploit semantically rich image features from a frozen DINOv2~\cite{oquab2023dinov2} backbone with two lightweight MLPs to perform semantic segmentation and object boundary. In a novel panoptic fusion module, we post-process the predictions with classical computer vision tools such as connected components analysis (CCA) and normalized cut (NCut) to obtain the merged panoptic output. A key characteristic of our proposed methods is that training them requires as few as ten annotated images. 
We further proposed to utilize our methods as a plug-in for existing methods. For instance, we trained our methods with ten annotated images from the Cityscapes dataset~\cite{cordts2016cityscapes} and subsequently produced pseudo-labels for a large set of unlabeled images. Due to the high quality of these pseudo-labels, one can train state-of-the-art panoptic segmentation models~\cite{cheng2022mask2former} that usually require large annotated datasets.
To conclude, we believe that our methods pose a highly impactful paradigm shift for deploying panoptic segmentation in the wild. 
}

{\parskip=3pt
\noindent\textbf{LiDAR-Based Mapping}
represents the third pillar of my previous research.
Fusing the complementary information from cameras and LiDAR~\cite{andresen2020accurate} typically requires a highly accurate calibration between them. Commonly, this is performed in a target-based manner, relying on artificial patterns and requiring human supervision. Although some effort has gone into automated calibration~\cite{guindel2017automatic, kim2019extrinsic}, even target-less methods often still need special data collection rendering prior works inefficient for frequent re-calibration and/or fleet application.
Thus, we introduced a novel method for automatic target-less camera-LiDAR calibration that requires neither human initialization nor special data recording~\cite{petek2024automatic}. We proposed to formulate extrinsic calibration as a graph optimization problem constrained by sensor motion and point correspondences. First, we match paths obtained from visual and LiDAR odometry. Then, we refine this initial registration with 2D pixel to 3D point correspondences from a neural network~\cite{cattaneo2024cmrnext}. The key novelty of our method is that in contrast to previous learning-based calibration approaches, we relax the black-box nature of the employed network by leveraging the estimated point correspondences in a classical optimization setup.

Finally, we address efficient robot learning via multi-agent collaboration~\cite{greve2024curb}. Although it has been shown that robotic systems can leverage high-definition map information as effective priors for several downstream tasks~\cite{cattaneo2024cmrnext, diazdiaz2022hd, yang2018hdnet}, such maps are often constructed via arduous labeling efforts. This is in contrast to our vision that modern mapping approaches should allow for frequent updates to capture structural changes and enable efficient access and information querying. Therefore, we proposed to utilize a hierarchical scene graph that incorporates multiple layers for roads, landmarks, and dynamic objects. While previous works have introduced similar concepts for indoor scenes~\cite{armeni20193d, hughes2022hydra}, our work is the first attempt towards constructing scene graphs for large-scale urban environments in a collaborative manner. In detail, we combine panoptic LiDAR readings from multiple agents in a central server that performs pose graph optimization incorporating both inter- and intra-agent loop closures to obtain a global panoptic 3D map. We then utilize this map to extract roads, intersections, and static landmarks. We further register other traffic participants observed by the ego agents to the lane graph allowing for inter-agent sharing of online detections.
}

\section{Ongoing and Future Directions}

Following these works, my ongoing research investigates how to transfer the insights gained from vision-based label-efficient panoptic segmentation to 3D point clouds. Specifically, we explore leveraging visual foundation models for processing LiDAR data, e.g., via transfer learning.
I am further extending the notion of efficiency to consider sensor cost. Addressing this area, we develop a mapping approach~\cite{schramm2023bevcar} that fuses surround view camera data with sparse radar measurements as an alternative to expensive LiDAR installations.

For future work, I intend to combine results from my previous projects to tie up loose ends in the final phase of my Ph.D. program. In detail, I am interested in merging our label-efficient segmentation techniques with robot mapping to obtain semantic representations of an environment without prior knowledge about present objects. These so-called open-world approaches are related to online continual learning in the sense that they can adapt to unforeseen situations. Finally, I will investigate the impact of foundation models on unsupervised domain adaptation. As demonstrated by prior research, they pose a powerful pretraining strategy and hence render a promising approach to mitigate forgetting.

\clearpage
\newpage




\bibliographystyle{plainnat}
\bibliography{references}


\end{document}